\documentclass[preprint,12pt]{elsarticle}

\usepackage{amssymb}
\usepackage{amsmath}

\usepackage{amssymb}
\usepackage{hyperref}
\usepackage{float}
\usepackage[symbol]{footmisc}

\graphicspath{{figures/}}

\journal{arXiv}

\begin{document}

\begin{frontmatter}



\title{SHAP-Based Supervised Clustering for Sample Classification and the Generalized Waterfall Plot}


\author{Justin Lin}
\affiliation{organization={Indiana University Mathematics Department},
            addressline={831 E 3rd St}, 
            city={Bloomington},
            postcode={47405}, 
            state={IN},
            country={USA}}
            
\author{Julia Fukuyama for the Alzheimer’s Disease Neuroimaging Initiative*}
\affiliation{organization={Indiana University Statistics Department},
            addressline={729 E 3rd St}, 
            city={Bloomington},
            postcode={47405}, 
            state={IN},
            country={USA}}

\begin{abstract}
In this growing age of data and technology, large black-box models are becoming the norm due to their ability to handle vast amounts of data and learn incredibly complex input-output relationships. The deficiency of these methods, however, is their inability to explain the prediction process, making them untrustworthy and their use precarious in high-stakes situations. SHapley Additive exPlanations (SHAP) analysis is an explainable AI method growing in popularity for its ability to explain model predictions in terms of the original features. For each sample and feature in the data set, we associate a SHAP value that quantifies the contribution of that feature to the prediction of that sample. Clustering these SHAP values can provide insight into the data by grouping samples that not only received the same prediction, but received the same prediction for similar reasons. In doing so, we map the various pathways through which distinct samples arrive at the same prediction. To showcase this methodology, we present a simulated experiment in addition to a case study in Alzheimer's disease using data from the Alzheimer’s Disease Neuroimaging Initiative (ADNI) database. We also present a novel generalization of the waterfall plot for multi-classification.
\end{abstract}

\begin{highlights}
\item SHAP-based supervised clustering provides insight undiscoverable in the raw data.
\item A SHAP-based classification explains model predictions in terms of the original features.
\item These explanations are necessary to understand and treat heterogeneous diseases like Alzheimer's disease.
\item In the context of multi-classification, the popular waterfall plot can be generalized to handle high-dimensional SHAP vectors.

\footnotetext[1]{Data used in preparation of this article were obtained from the Alzheimer’s Disease Neuroimaging Initiative (ADNI) database (adni.loni.usc.edu). As such, the investigators within the ADNI contributed to the design and implementation of ADNI and/or provided data but did not participate in analysis or writing of this report. A complete listing of ADNI investigators can be found at: http://adni.loni.usc.edu/wp-content/uploads/how\_to\_apply/ADNI\_Acknowledgement\_List.pdf}
\end{highlights}


\end{frontmatter}



\section{Introduction}

Supervised learning is a machine learning paradigm in which the goal is to predict outputs given known input-output pairs. Examples include regression, in which the output is one-dimensional and continuous, or classification, in which the output is nominal. Given the advances in data collection and computing power, increasingly complex methods are being developed to handle supervised data, many of which are known as black-box methods. As the name suggests, the inner workings of these methods are nebulous due to their high levels of complexity. However, they are among the highest performers when it comes to supervised learning.

In efforts to dissect and better understand these methods, explainable AI (XAI) methods are being developed to extract meaning from these so-called black-box models. One of which, SHapley Additive exPlanations (SHAP) analysis, is growing in popularity because of its ability to explain model predictions in terms of the original features. By linking predictions back to the original features, we can better understand why each sample receives the prediction it does and draw inference from these explanations.

For each sample and feature in the data, we calculate a corresponding SHAP value that quantifies the contribution of that feature to the prediction of that sample. These SHAP values can be organized in a matrix with dimensions equal to the original data set. However, the information the SHAP values encode is unique and highly valuable. A clustering of SHAP values would correspond to a classification of samples according model explanation. Samples belonging to the same cluster would not only have similar predictions, they would also have received similar predictions for similar reasons. By clustering SHAP values, as opposed to the original data, we are mapping out the various pathways through which distinct samples can arrive at similar predictions. In the context of medicine, this can help explain disease heterogeneity and progress precision medicine.

There is limited literature on the clustering SHAP values \cite{ex1, ex2, ex3}. In this paper, we hope to familiarize more scientists with the idea and showcase its application to the field of medicine, specifically. We present a controlled, simulated experiment, as well as a case study in Alzheimer's disease. We also present a novel generalization of the waterfall plot for multi-classification problems.

\section{Background}

\subsection{Supervised Clustering Workflow}

The supervised clustering methodology consists of five steps – predictive modeling, SHAP analysis, visualization, cluster analysis, and cluster interpretation (Figure 1). The exact methods used in some steps are interchangeable and can be adapted to the specific use case. In our experiments, we use eXtreme Gradient Boosting (XGBoost) for predictive modeling, Uniform Manifold Approximation and Projection (UMAP) for visualization, and Hierarchical Density-Based Spatial Cluster of Applications with Noise (HDBSCAN) for cluster analysis.

\begin{figure}[H]
\centering
\includegraphics[width=4in]{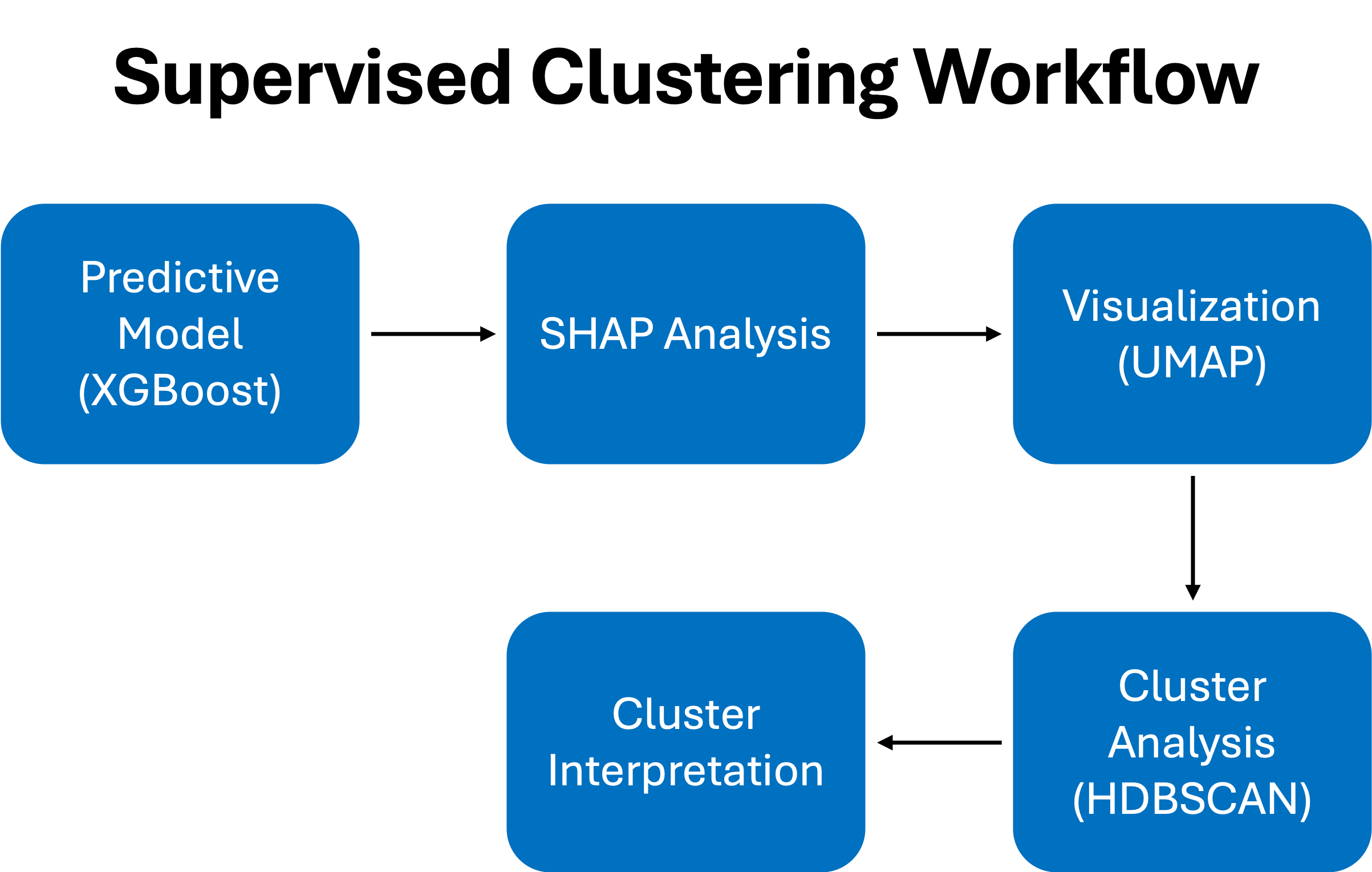}
\caption{Supervised clustering workflow. The methods we used within in each step are parenthesized.}
\end{figure}
 
This methodology allows us to categorize samples according to model explanation, which is often more insightful than the raw data. This categorization groups together samples that are predicted to be in the same target class for similar reasons. In the context of medicine, for example, this process can explain various disease trajectories by subgrouping both healthy and unhealthy patients according to model explanation. Patients with the same disease can be predicted to be unhealthy for completely distinct reasons, especially when dealing with heterogeneous diseases. A better understanding of disease heterogeneity leads to individualized treatment that is more efficacious than catch-all treatment.

Following the cluster analysis, examination of the raw data provides interpretation of the categorization. Heatmaps and waterfall plots can be used to analyze the allocation of each prediction across the features for different samples. The SHAP values reveal which features drive prediction, while the raw values reveal the ways in which they drive prediction.

The consideration of model explanation, instead of the raw data alone, is necessary to study the input-output relationships hidden in the data. Predictive modeling approximates this highly complex function, and SHAP analysis quantifies these relationships in terms of the original features, providing means for analysis. Through this procedure, we are able to produce a data-driven classification of samples, undiscoverable in the raw data, that describes the various sample archetypes belonging to each target class.

\subsection{XGBoost}

XGBoost, or eXtreme Gradient Boosting, is an algorithm based on gradient boosted decision trees developed to handle large and complex data sets efficiently. The concept behind gradient boosted decision trees is to sequentially train new decision trees that predict the residuals of the previous ensemble of trees. In doing so, each tree is correcting the errors of the previous ensemble of trees and polishing the complete model. The final ensemble of trees can then be used to accurately predict the output of new samples. Note, XGBoost can be applied to both regression and classification problems. See \cite{xgboost} for details.

\subsection{SHAP Analysis}

SHAP analysis provides a representation of the learned relationships in terms of the original features. Given a sample, each feature is assigned an associated SHAP value quantifying the contribution of that feature towards the prediction of the given sample. Thus, for each sample, its prediction is allocated among the individual features, indicating the magnitude and direction of contribution of each feature. Unlike covariate analysis, SHAP analysis is conducted on a sample-by-sample basis. For one feature, there are individual SHAP values associated with each sample. In other words, the effect of each covariate varies sample to sample.

To provide a mathematical formulation, suppose $f:X \to \mathbb{R}$ is a trained model, where $f$ is trained on a data set $X' \subset X \subset \mathbb{R}^p$. In the case of a binary target variable, the model predicts the log-odds of success. For each $x \in X$, there exists a sequence of SHAP values $\phi(f;x)_1, \hdots, \phi(f;x)_p$, one associated to each feature, such that
\begin{equation*}
\sum_{i=1}^p \phi(f;x)_i = f(x) - \mathbb{E}[f(X')].
\end{equation*}
In other words, the SHAP values for a given sample sum to the difference between the output of that sample and the average output of the entire training set. The deviation of the sample’s prediction from the average prediction is allocated among the individual features.

It is important to note each SHAP value $\phi(f;x)_i$ can be positive or negative. When predicting log-odds, negative SHAP values contribute towards a prediction of failure, while positive SHAP values contribute towards a prediction of success. The significance of each feature in deciding the prediction is then proportional to the magnitude of the corresponding SHAP value.

These SHAP values are based on Shapley values, originally coined by Lloyd Shapley in 1953 as way to quantify the contribution of each player to the outcome of a game \cite{shapley}. Developed by Scott M. Lundberg and Su-In Lee \cite{shap}, SHAP analysis connects Shapley values to machine learning by treating each feature as a player and the target variable as the outcome of the game. The exact estimation of the SHAP values is dependent on the predictive model. See \cite{treeshap} for more information on TreeSHAP, the algorithm used to estimate SHAP values for tree-based models like XGBoost.

In the case of multi-classification, the SHAP values are $k$-dimensional vectors, where $k$ is the total number of classes. Each component of a SHAP vector represents the contribution of the corresponding feature to the prediction of the corresponding class. Hence, the SHAP vectors can be organized in a $(n \times p \times k)$-dimensional tensor, where $n$ is the number of samples, $p$ is the number of features, and $k$ is the number of classes. To analyze, we can flatten the last two dimensions to form a $(n \times pk)$-dimensional matrix.

The SHAP values are calculated using a repeated cross-validation process. First, the data is partitioned into a series of $l$ folds. The SHAP values for each fold are calculated upon a model trained on the other $l-1$ folds. This out-of-sample calculation ensures the SHAP values are not overfitting the noise. This process is then repeated and the SHAP values are averaged across runs to eliminate bias introduced by the random partitioning.

\subsection{UMAP}

UMAP belongs to a family of dimension reduction methods whose goal is to represent a high-dimensional data set using a fewer number of features. Being able to represent a data set using only two features allows us to plot the data on a two-dimensional set of axes for visualization. UMAP is the state-of-the-art dimension reduction method for visualization and cluster detection because of its prioritization of local structure. Its nonlinear nature also allows it to handle large reductions in the number of dimensions.  See \cite{umap} for details.

\subsection{HDBSCAN}

HDBSCAN is a hierarchical clustering method capable of handling noise and clusters with variable densities. Its predecessor, DBSCAN, is a flat clustering method that struggles with clusters of variable densities just like KMeans. HDBSCAN solves this issue by implementing a hierarchical procedure at the cost of computational efficiency. See \cite{hdbscan} for details.

\section{Methods}

\subsection{Data}

\subsubsection{Simulated Data}

To emulate a multi-classification problem, we assume the data follow a multinomial logistic regression model. In particular, the probabilities of belonging to each of three classes are linked to independent linear combinations of the input variables via the logistic function. The coefficients of each linear combination are chosen to create a scenario in which there exist distinct pathways to the same target class. 

The coefficients are defined as follows,
$$f_1(x_1, \hdots, x_{10}) = 4x_1x_2 + 4x_1 + 4x_2 + \sum_{i=3}^{10} \beta_{1,i}x_i$$
$$f_2(x_1, \hdots, x_{10}) = 4x_1x_2 - 4x_1 - 4x_2 + \sum_{i=3}^{10} \beta_{2,i}x_i$$
where $\beta_{j,i} \overset{iid}{\sim} N(0, 1)$. With these coefficients, we expect Class 1 to contain the samples whose first two entries are positive and Class 2 to contain the samples whose first two entries are negative. The remainder of the samples, those whose first two entries are opposite signs, should belong to Class 3. There are two distinct pathways to arrive at Class 3: $x_1 > 0, x_2 < 0$ and $x_1 < 0, x_2 > 0$.

The input data are uniformly sampled from a hyperrectangle,
\begin{equation*}
X_1, \hdots, X_{1500} \overset{\textrm{iid}}{\sim} \textrm{Unif}([-5, 5]^{10}).
\end{equation*}
We sample 1,500 points in 10 dimensions. Then for each sample point $X_i$, we calculate the probabilities of belong to each class
$$p_{1,i} = \frac{\exp(f_1(X_i))}{1 + \exp(f_1(X_i)) + \exp(f_2(X_i))}$$
$$p_{2,i} = \frac{\exp(f_2(X_i))}{1 + \exp(f_1(X_i)) + \exp(f_2(X_i))}$$
$$p_{3,i} = \frac{1}{1 + \exp(f_1(X_i)) + \exp(f_2(X_i))}$$
Lastly, the final class labels are sampled from multinomial distributions with the associated probabilities,
\begin{equation*}
y_i \sim \textrm{Multinom}([p_{1,i}, p_{2,i}, p_{3,i}]).
\end{equation*}

\subsubsection{ADNI Data}

The Alzheimer's Disease Neuroimaging Initiative (ADNI) maintains a collection of longitudinal clinical, imaging, genetic, and other biomarker data. The ADNI was launched in 2003 as a public-private partnership, led by Principal Investigator Michael W. Weiner, MD. The primary goal of ADNI has been to test whether serial magnetic resonance imaging (MRI), positron emission tomography (PET), other biological markers, and clinical and neuropsychological assessment can be combined to measure the progression of mild cognitive impairment and early Alzheimer’s disease. For up-to-date information, see www.adni-info.org.

The various Alzheimer's data sets were combined using the ADNIMERGE R package \cite{adnimerge}. Visit data (visit code, exam date, site, study protocol), categorial demographic information (sex, ethnicity, marriage status), and equipment details (MRI Field Strength, FreeSurfer Software Edition, LONI image ID) were removed. The features missing readings for more than half of the patients were also removed (seven in total).

The cleaned data set contains 2,422 patients and 39 features. Each column was linearly re-scaled to range from zero to one to account for varying units of measurement. The target variable is comprised of three classes -- Cognitively Normal (CN), Mild Cognitive Impairment (MCI), and Alzheimer's/Dementia (AD).

\subsection{High-Dimensional Waterfall Plots}

Waterfall plots are designed to display explanations for individual samples. Given a sample, the features are ordered according to the absolute SHAP values and bars are used to represent each SHAP value. However, the base of each bar is located at the end of the previous bar so that the bars represent the cumulative sum of the SHAP values. When anchored at the average prediction, the series of bars ends at the prediction of the given sample. Hence, the waterfall plot describes a path from the average prediction to the given sample's prediction.

In the case of multi-classification, however, the SHAP vectors do not give way to a waterfall plot. Bars can only illustrate one-dimensional SHAP values. To generalize the waterfall plot to multi-classification, we can represent the explanation of a sample with a path in $k$ dimensions. Each segment of the path corresponds to one feature. These high-dimensional paths can then be visualized in one of two ways. The first way is to project the paths onto the subspace spanned by two of the classes, or equivalently, ignore the SHAP values corresponding to every other class. These projections are ideal for making pairwise comparisons between the two chosen classes. The second way is to project the paths onto the two-dimensional subspace that retains the most information using Principal Component Analysis (PCA). This projection retains information about all $k$ classes, but makes the axes slightly more difficult to interpret. The axes no longer represent individual classes, but a linear combination of them. Further techniques, like the biplot, are then needed to interpret the axes.

\subsection{Code Availability}

Code for the simulated experiment is freely available at \url{https://github.com/JustinMLin/SHAP\_clustering}.

\section{Results}

\subsection{Simulated Data}

When visualizing the raw data with UMAP, there are no meaningful clusters or trends (Figure 2).

\begin{figure}[H]
\centering
\includegraphics[width=4in]{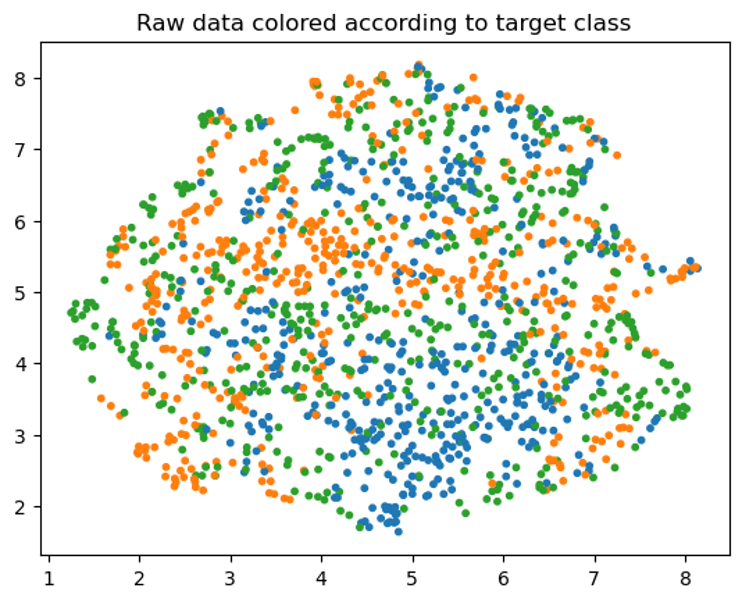}
\caption{Raw data embedded in two dimensions with UMAP and colored according to target class.}
\end{figure}

The trained XGBoost model exhibits great performance on an out-of-sample test set (Table 1), indicating the model explanations are reliable.

\begin{table}[H]
\centering
\begin{tabular}{| c | c c c c |}
 \hline
  Class & Precision & Recall & F1-score & Support\\
  \hline
  0 & 0.92 & 0.93 & 0.92 & 156\\
  1 & 0.89 & 0.93 & 0.91 & 135\\
  2 & 0.90 & 0.86 & 0.88 & 159\\
  \hline\hline
  accuracy & & & 0.90 & 450\\
  macro avg & 0.90 & 0.90 & 0.90 & 450\\
  weighted avg & 0.90 & 0.90 & 0.90 & 450\\
  \hline
\end{tabular}
\caption{Performance metrics for XGBoost model on simulated data.}
\end{table}

Figure 3 illustrates the average absolute SHAP values for each feature, revealing the most influential features on average. Unsurprisingly, Features 1 and 0 are the largest contributors to prediction.

\begin{figure}[H]
\centering
\includegraphics[width=4in]{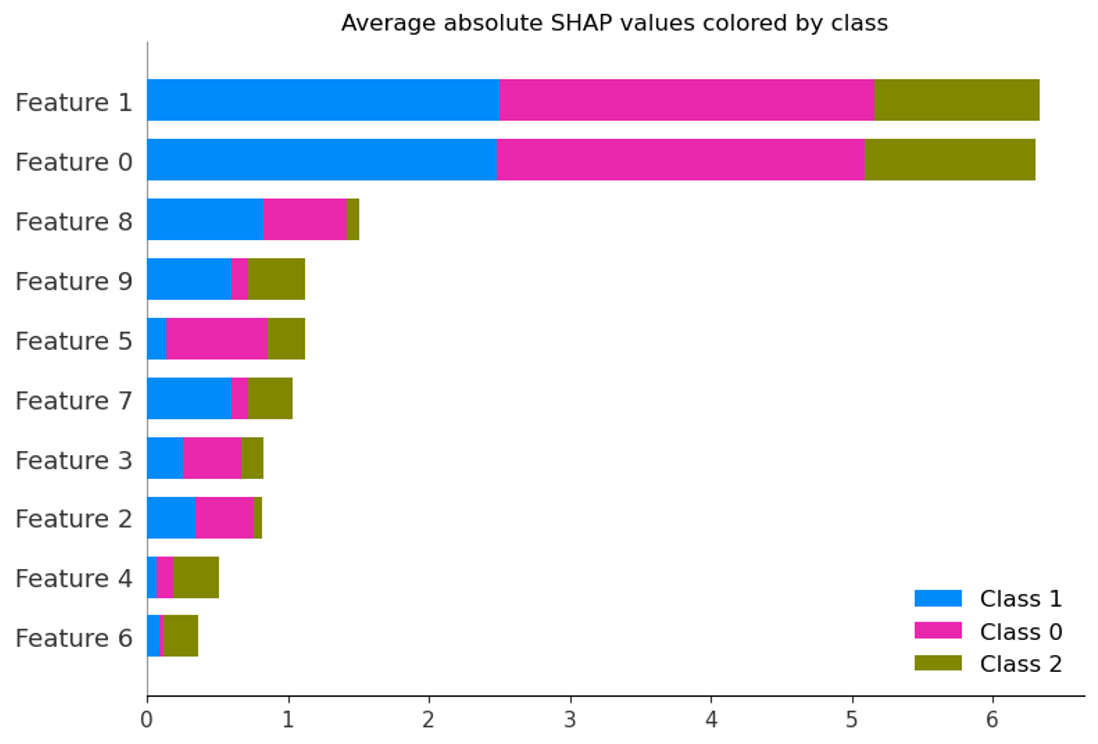}
\caption{Absolute response function coefficients and average absolute SHAP values.}
\end{figure}

In comparison to the raw data, the SHAP values tell a much more meaningful story. Figure 4 shows both the SHAP values and raw data colored according to an HDBSCAN clustering of the SHAP values. There are four distinct clusters that represent samples with similar predictions and model explanations.

\begin{figure}[H]
\centering
\includegraphics[width=4in]{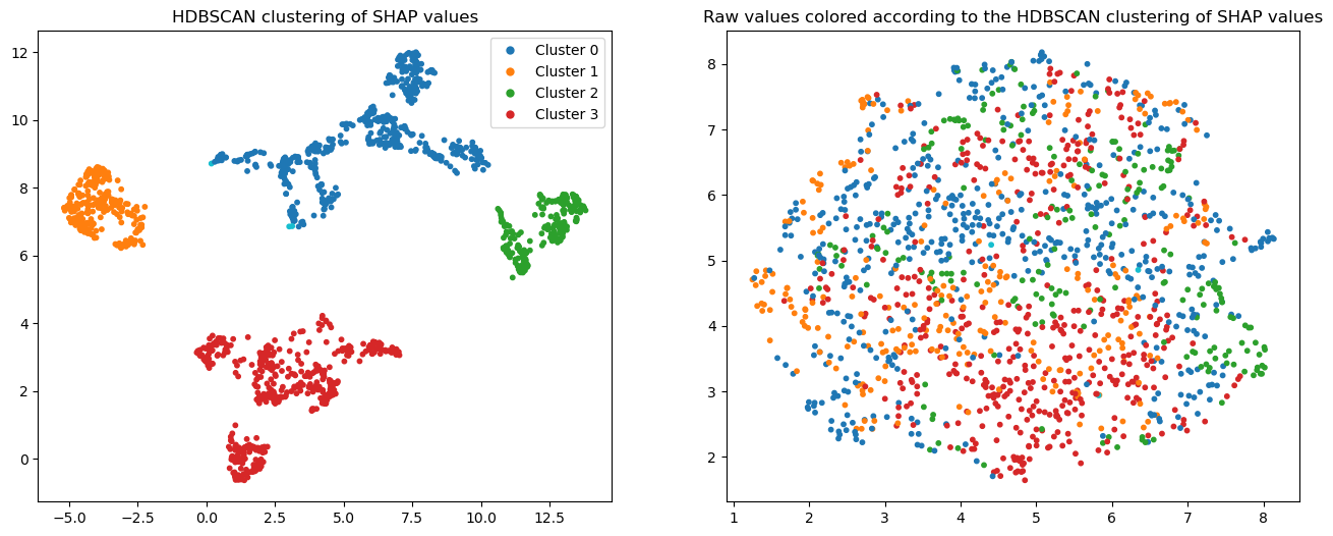}
\caption{HDBSCAN clustering of the SHAP values (left). Raw values colored according to the same clustering (right).}
\end{figure}

To interpret these clusters, we use a high-dimensional generalization of the waterfall plot. By treating the average SHAP vectors as paths in high-dimension, we create feature-by-feature path representations of each cluster. A projection of these paths onto two dimensions gives way for visualization (Figure 5). 

\begin{figure}[H]
\centering
\includegraphics[width=4in]{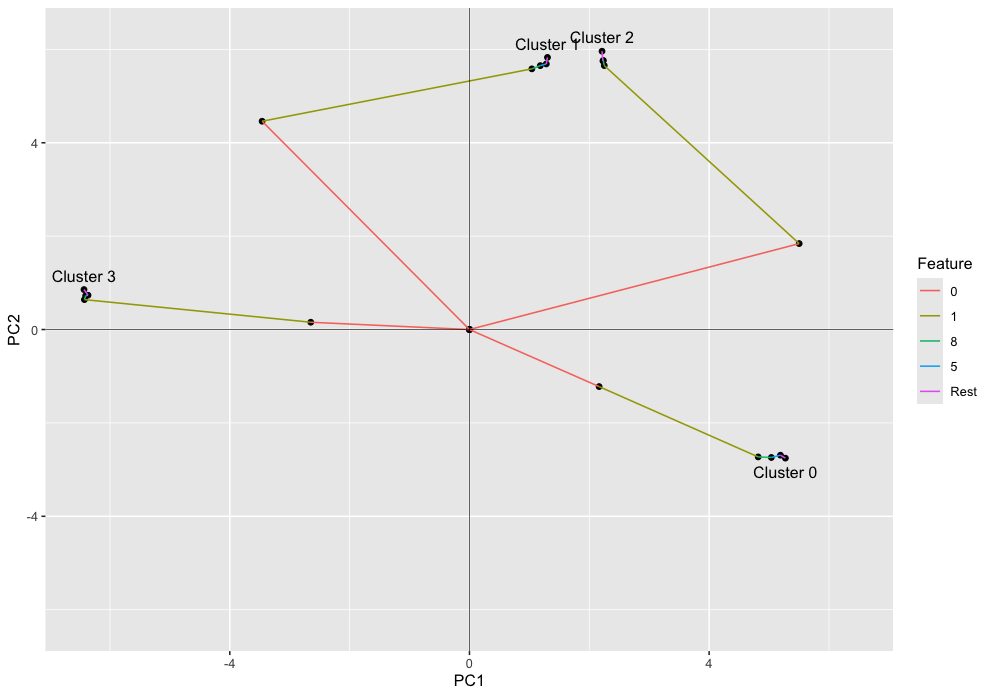}
\caption{Waterfall plot of top SHAP values averaged across clusters.}
\end{figure}

The paths can be partitioned into three separate groups: Cluster 0, Cluster 3, and Clusters 1/2. Clusters 1 and 2 illustrate how multiple samples can arrive at the same prediction through distinct pathways. This phenomenon was only made apparent by incorporating a predictive model into the usual cluster analysis workflow. Without model explanations, there would be no distinguishing between subgroups.

If we view the SHAP values colored according to target class, we see the three groups pinpointed in the generalized waterfall plot correspond to the three target classes (Figure 6).

\begin{figure}[H]
\centering
\includegraphics[width=4in]{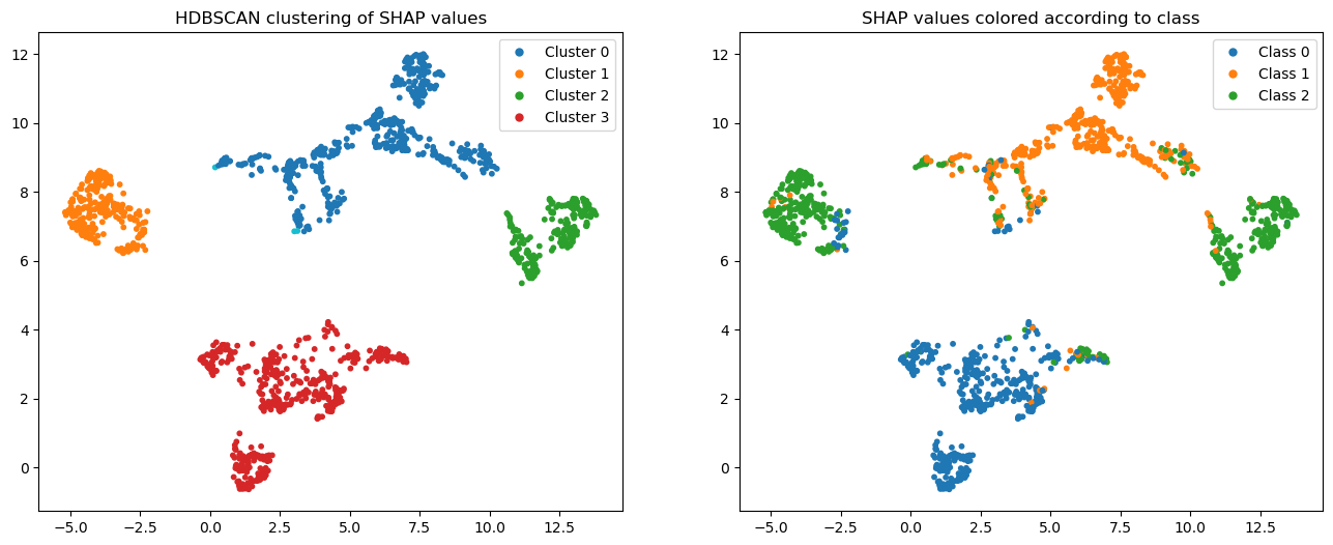}
\caption{HDBSCAN clustering of the SHAP values (left). SHAP values colored according to target class (right).}
\end{figure}

Furthermore, the clusters correspond to the parities of Features 0 and 1 (Figure 7). Cluster 3 contains the samples whose values for Features 0 and 1 are both positive. Cluster 0 contains the samples whose values for Features 0 and 1 are both negative. And Clusters 1/2 contain the samples whose values for Features 0 and 1 are opposite signs. The supervised clustering workflow was able to learn the function used to label the data and provide a corresponding classification of samples.

\begin{figure}[H]
\centering
\includegraphics[width=4in]{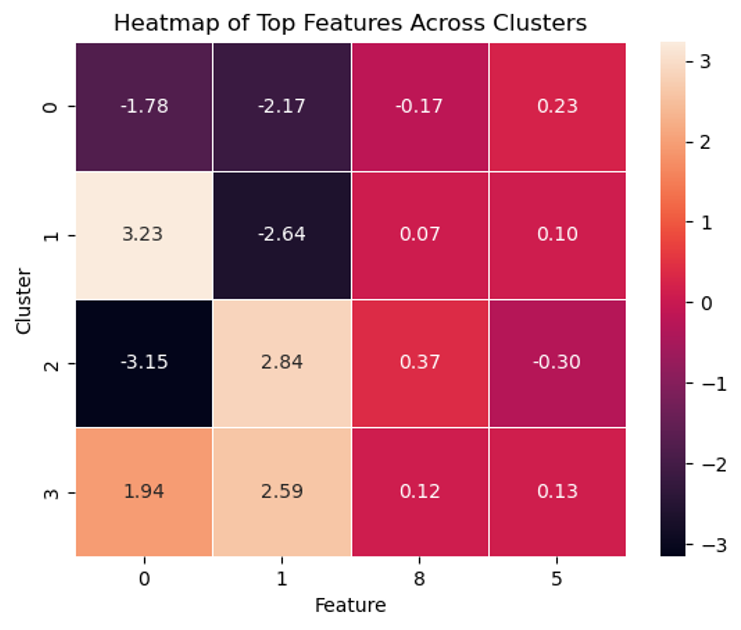}
\caption{Heatmap of raw data averaged across clusters.}
\end{figure}

This simulated example demonstrates the SHAP-based clustering's ability to further classify samples beyond target class. Class 2 was discovered to contain two distinct subgroups of samples -- those with positive values for Feature 0 and negative values for Feature 1, and those with negative values for Feature 0 and positive values for Feature 1. While this situation is purely hypothetical, one can imagine the value of a finer classification in various real-world situations.

\subsubsection{Finer Clustering}
One can argue Cluster 3 is, in fact, two distinct clusters (Figure 8). If we repeat the analysis with a finer clustering, we see the distinction between Clusters 3 and 4.

\begin{figure}[H]
\centering
\includegraphics[width=4in]{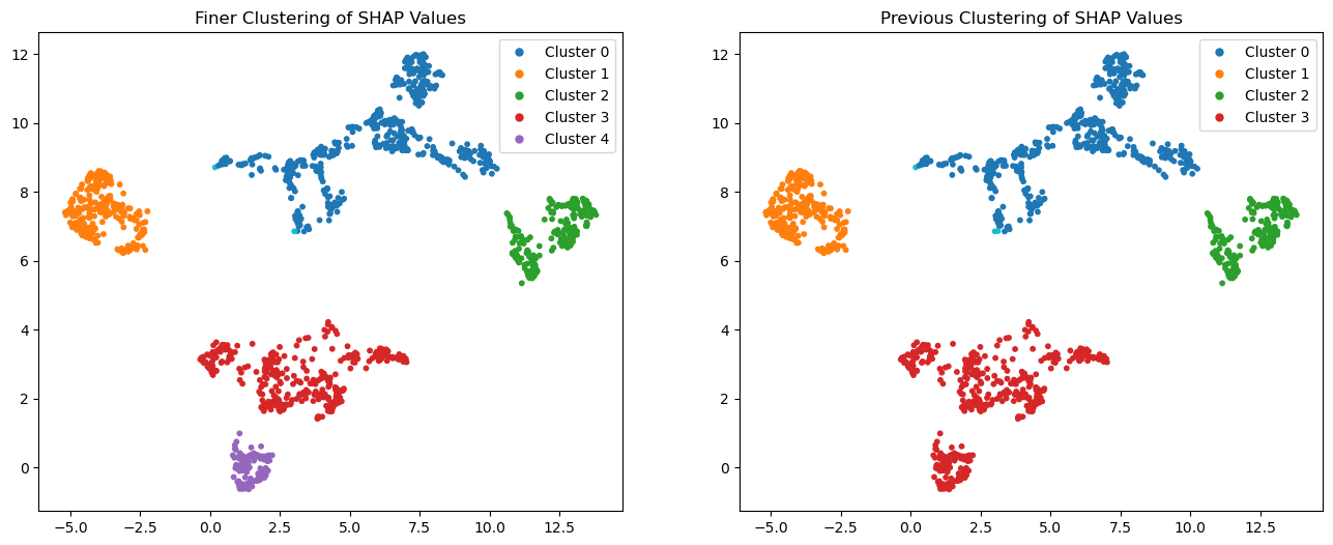}
\caption{Finer clustering of SHAP values (left) compared to previous clustering (right).}
\end{figure}

The waterfall plot depicts Clusters 3 and 4 having very similar model explanations (Figure 9). The only minor difference is Feature 8, as demonstrated by the opposite directions of the segments corresponding to Feature 8 in each path.

\begin{figure}[H]
\centering
\includegraphics[width=4in]{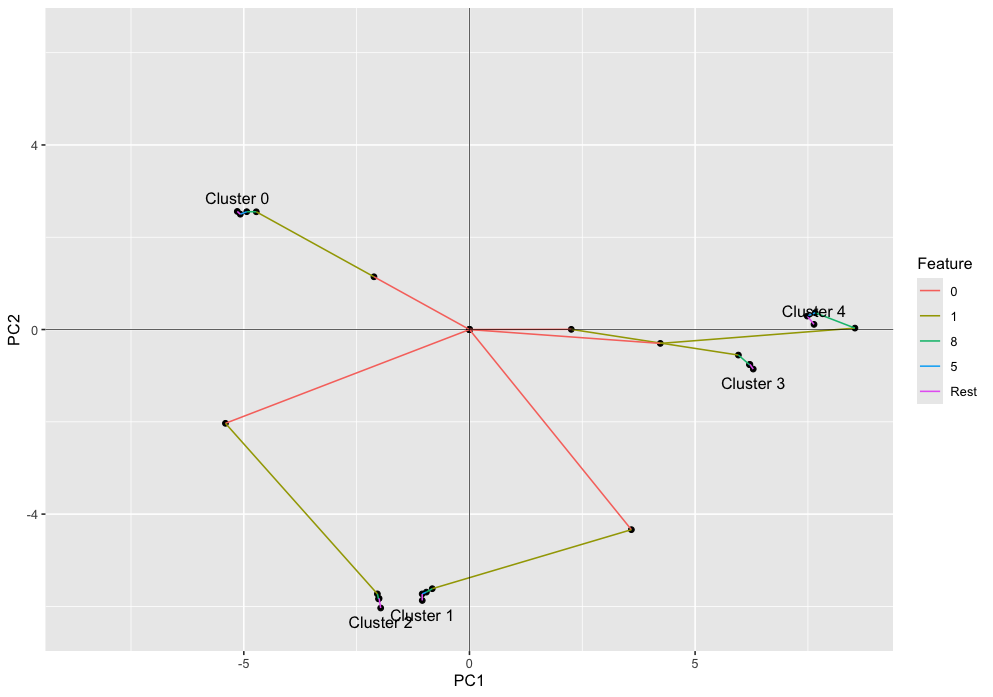}
\caption{Waterfall plot of top SHAP values averaged across finer clusters.}
\end{figure}

This hypothesis is confirmed by the heatmap of average raw values (Figure 10). Both Clusters 3 and 4 are positive in Features 0 and 1, but they differ in Feature 8. Cluster 4 is defined by large negative values for Feature 8, while Cluster 3 is slightly positive, on average, for Feature 8.

\begin{figure}[H]
\centering
\includegraphics[width=4in]{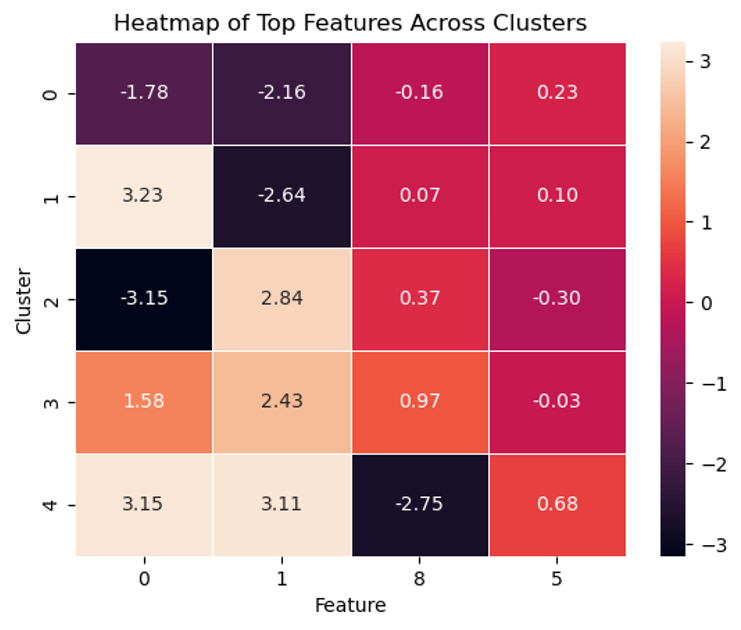}
\caption{Heatmap of raw data averaged across finer clusters.}
\end{figure}

This finer clustering demonstrates the stability of SHAP values and the high-dimensional waterfall plot. The methodology was able to detect a difference in Feature 8, a less impactful feature by design, without obfuscating the signal. Both the embedding of the SHAP values and the high-dimensional waterfall plot primarily illustrated four major clusters, corresponding to the underlying classes of the simulated data.

\subsection{ADNI Data}

The raw data does not exhibit any clustering, but the different patient statuses are certainly segregated within the data (Figure 11 left). There exist two distinct subgroups of cognitively normal patients, which correspond to the number of APOE e4 alleles, a variation of the APOE (apolipoprotein E) gene known to correlate with the development of Alzherimer's (Figure 8 right).

\begin{figure}[H]
\centering
\includegraphics[width=4in]{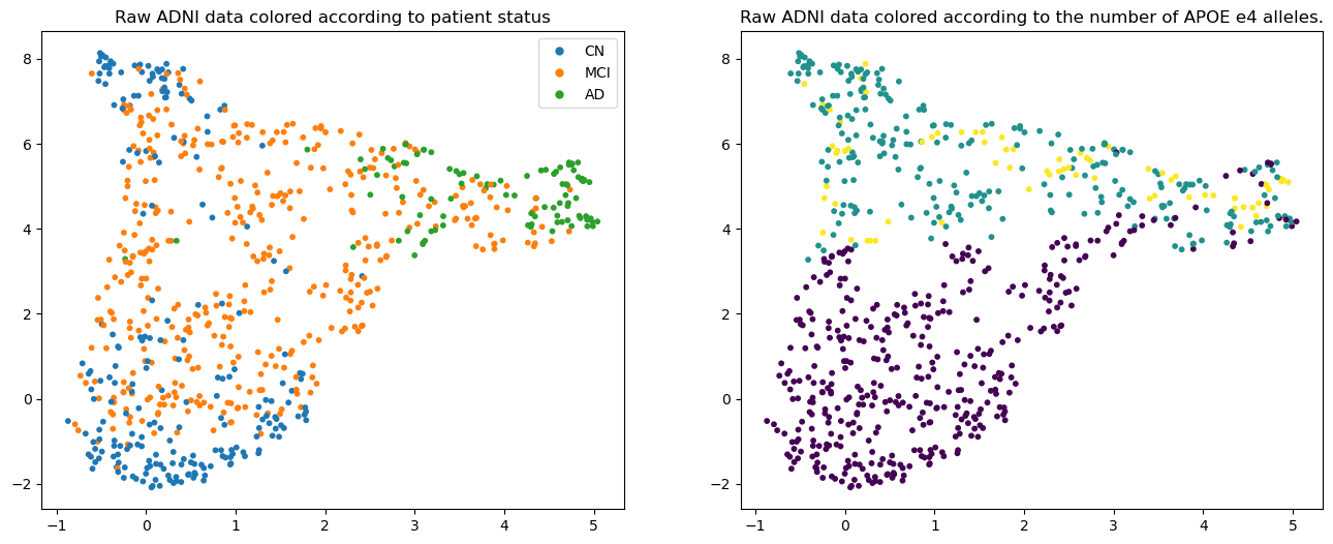}
\caption{Raw data embedded in two dimensions with UMAP colored according to patient status (left) and APOE4 (right).}
\end{figure}

The trained XGBoost model exhibits adequate performance on an out-of-sample test set (Table 2). Note the uneven class sizes bias the performance metrics, but performance on the smallest class is still sufficient.

\begin{table}[H]
\centering
\begin{tabular}{| c | c c c c |}
 \hline
  Class & Precision & Recall & F1-score & Support\\
  \hline
  CN & 0.98 & 0.95 & 0.96 & 268\\
  MCI & 0.91 & 0.93 & 0.92 & 336\\
  AD & 0.87 & 0.85 & 0.86 & 123\\
  \hline\hline
  accuracy & & & 0.93 & 727\\
  macro avg & 0.92 & 0.91 & 0.91 & 727\\
  weighted avg & 0.93 & 0.93 & 0.93 & 727\\
  \hline
\end{tabular}
\caption{Performance metrics for XGBoost model on ADNI data.}
\end{table}

Because this is a multi-classification model, the SHAP values are three-dimensional vectors. In particular, there are three distinct sets of SHAP values quantifying the contributions of each feature to each of the three target classes. Figure 12 summarizes the average contribution of each feature across all three classes. CDRSB is the dominating contributor, particularly to the CN and MCI classes. The Clinical Dementia Rating (CDR, \cite{cdrsb}) is a measure of cognition obtained by interviewing both the patient and a care partner. The questionnaire measures six different domains including memory and problem-solving. CDRSB is the sum of boxes, or sum of scores, across the six domains.

\begin{figure}[H]
\centering
\includegraphics[width=4in]{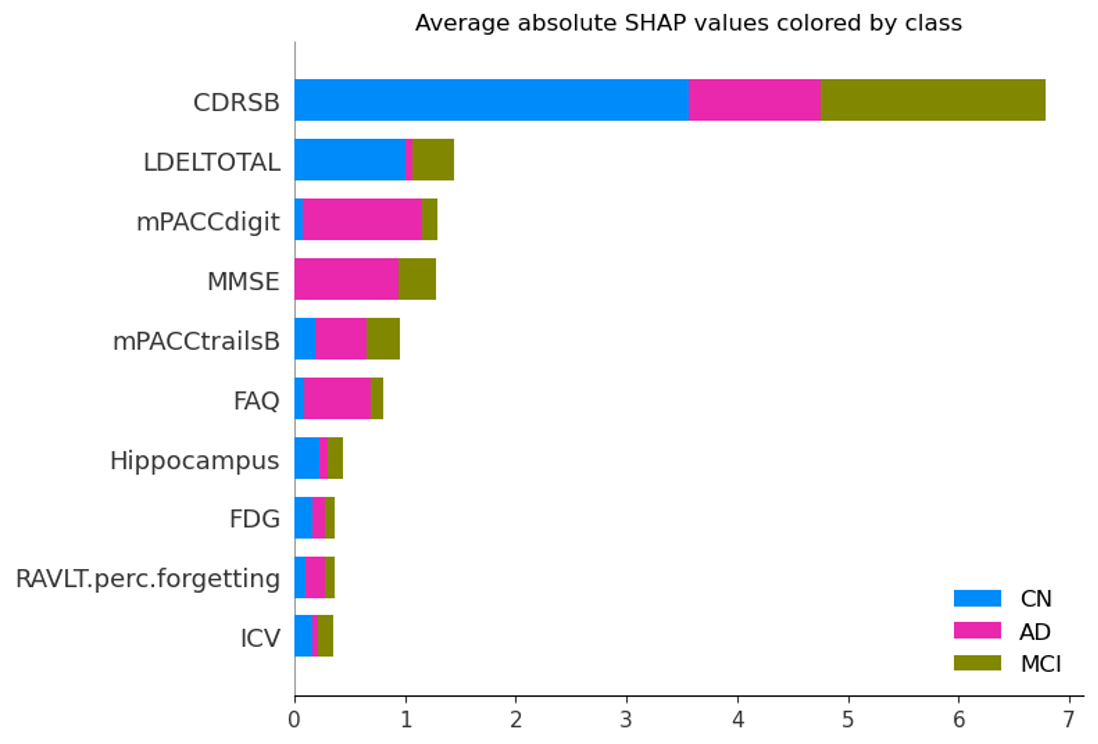}
\caption{Average absolute SHAP values colored by class.}
\end{figure}

The following three features, LDELTOTAL, mPACCdigit, and MMSE, also play non-trivial roles in classification, especially in pairwise comparisons. LDELTOTAL plays a large role in distinguishing CN and MCI patients, while mPACCdigit and MMSE help distinguish MCI and AD patients. LDELTOTAL is one of the scores in the Logical Memory - Delayed Recall portion of the Luria-Nebraska Neuropsychological Assessment Battery, a standardized test used to measure cognitive skills and functions. mPACCdigit is the score to the Preclinical Alzheimer's Cognitive Composite (PACC) test. MMSE is the score to the Mini Mental State Examination.

To analyze the SHAP values on a sample-by-sample basis, we embed the flattened SHAP vectors using UMAP (Figure 13). Unlike the raw data, the SHAP values cluster into multiple distinct groups. Even within patients statuses, there exist subgroups.

\begin{figure}[H]
\centering
\includegraphics[width=4in]{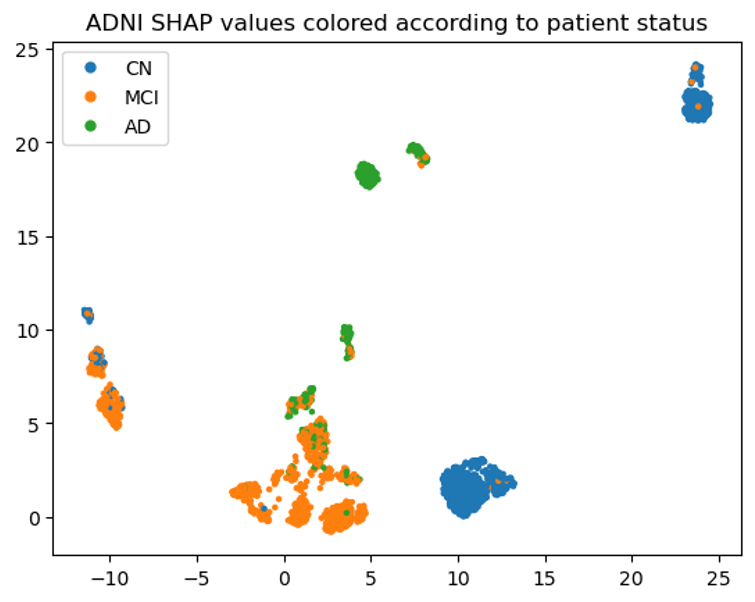}
\caption{ADNI SHAP values embedded with UMAP and colored according to patient status.}
\end{figure}

To study these clusters, we conduct a cluster analysis using HDBSCAN (Figure 14). Note the two CN subgroups, separated according to APOE e4, belong to the same cluster of SHAP values (purple in Figure 11). This implies the APOE4 feature contributes similarly to prediction in both subgroups of CN patients, despite having differing numbers of APOE e4 alleles. This is consistent with the relationship between the APOE4 feature and target class. Figures 7 and 8 show higher APOE4 values correlate only with the AD class. Within the CN and MCI classes, there exist patients with both low and high APOE4 values.

\begin{figure}[H]
\centering
\includegraphics[width=4in]{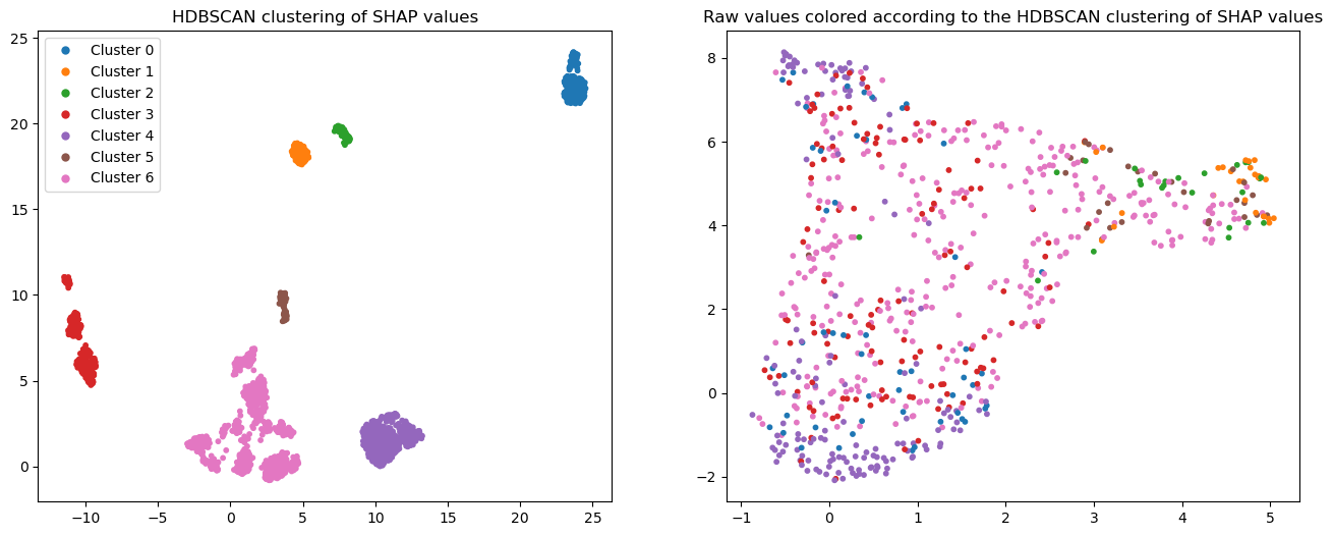}
\caption{HDBSCAN clustering of the SHAP values (left). Raw values colored according to the same clustering (right).}
\end{figure}

To interpret these clusters, we use the generalized waterfall plot. Figure 15 show the projection of the high-dimensional SHAP paths averaged across clusters. The paths can be classified into three separate groups -- Clusters 0 and 4 correspond to CN patients, Clusters 3 and 6 correspond to MCI patients, and Clusters 1, 2, and 5 correspond to AD patients. Moreover, Cluster 3 leans more towards the CN group and Cluster 6 leans more towards the AD group, which is consistent with Figure 10.

\begin{figure}[H]
\centering
\includegraphics[width=4in]{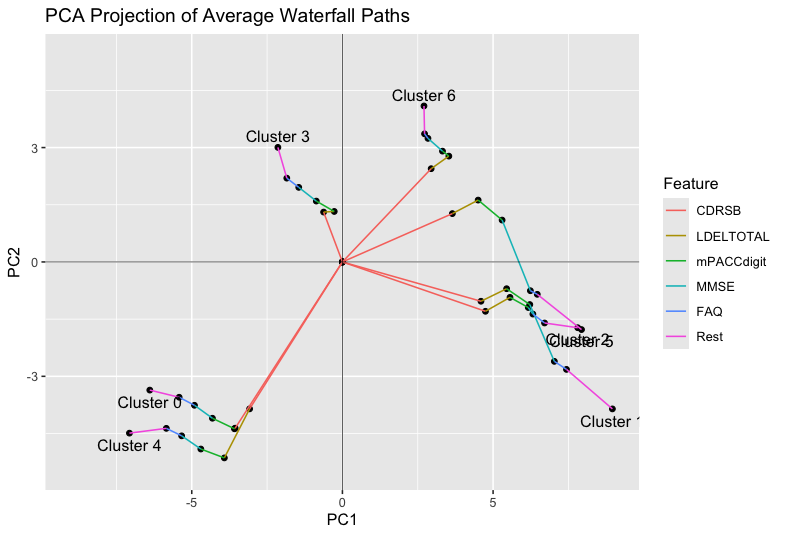}
\caption{PCA projection of high-dimensional waterfall paths averaged across HDBSCAN clusters.}
\end{figure}

The paths illustrate the various pathways through which different patients arrive at the same prediction. For example, Clusters 2 and 5 have almost identical predictions on average, but the CDRSB feature contributes differently on average between the two groups of patients. For Cluster 2, the average SHAP value for CDRSB towards AD is 1.91, while the corresponding average for Cluster 5 is 0.57. According to the paths, the lack of predictive power of CDRSB in Cluster 5 is made up by MMSE. The average SHAP value for MMSE towards AD is 0.22 in Cluster 2 and 1.21 in Cluster 5.

CDRSB also seems to be the differentiating feature between Clusters 3 and 6. While both clusters correspond to MCI  patients, Cluster 3 also contains a small proportion of CN patients and Cluster 6 contains a small proportion of AD patients. The SHAP values (Table 3) indicate CDRSB is a strong predictor against AD in Cluster 3 and a strong predictor against CN in Cluster 6. This suggests the CDR evaluation is a good indicator of whether those with mild cognitive impairment are at risk of Alzheimer's disease because it can be used to place MCI patients into Cluster 3 or 6.

\begin{table}[H]
\centering
\begin{tabular}{| c | c c c |}
 \hline
  Cluster & CN Mean & MCI Mean & AD Mean \\
  \hline
  3 & -0.52 & 0.92 & -1.50\\
  6 & -4.31 & 1.19 & -0.50\\
  \hline
\end{tabular}
\caption{Average SHAP values for CDRSB in Clusters 3 and 6.}
\end{table}

Through supervised clustering of SHAP values, we've shown the trichotomy of cognitively normal, mild cognitive impairment, and Alzheimer's/Dementia is more nuanced than the raw data suggests. Within each class, there exist subgroups that arrive at the same prediction via distinct pathways. This distinction can be very helpful in the diagnosis and treatment of Alzheimer's disease. For example, Clusters 2 and 5 both correspond to AD patients, but the distinct driving factors, CDRSB and MMSE, may test different aspects of cognitive ability. In particular, the optimal treatment plan may differ between these two groups of patients.

\section{Discussion}

Through simulation and a case study in Alzheimer's, we have showcased the ability of SHAP-based clustering to reveal insightful classifications undiscoverable in the raw data. In particular, the methodology approximates the response function, which is often the primary component of study, and clusters accordingly. For example, the SHAP values in our simulation experiment clustered according to the logistic regression model coefficients, describing how classes are assigned in terms of the original features. In practice, this function is incredibly complex and unknown, but even an interpretable approximation is invaluable.

The Alzheimer's case study was a demonstration of a real use case. Machine learning methods in Alzheimer's, and on the ADNI data specifically, has been well-documented \cite{ML_adni1, ML_adni2}, but previous methods lack actionable interpretation. A SHAP-based clustering provides much needed context for any black-box model by explaining the different avenues of prediction in terms of the original features. Our novel generalization of waterfall plots allows scientists to visualize these avenues, providing interpretation of the clusters. Such interpretations can be leveraged to make data-driven decisions with disease and patient heterogeneity in mind. Pinpointing patient archetypes and mapping the prediction of each archetypical patient can further precision medicine and treatment of heterogeneous diseases like Alzheimer's.

Further works should explore high-dimensional generalizations of other SHAP analysis plots, including bar plots, heatmaps, beeswarm plots, violin plots, etc. With more than two classes, one can always draw a plot for each class separately, but such a collection is difficult to interpret and ignores crucial interactions between classes. This is especially problematic when the number of classes is more than a few.

\section{Acknowledgements}
Data collection and sharing for this project was funded by the Alzheimer's Disease Neuroimaging Initiative (ADNI) (National Institutes of Health Grant U01 AG024904) and DOD ADNI (Department of Defense award number W81XWH-12-2-0012). ADNI is funded by the National Institute on Aging, the National Institute of Biomedical Imaging and Bioengineering, and through generous contributions from the following: AbbVie, Alzheimer’s Association; Alzheimer’s Drug Discovery Foundation; Araclon Biotech; BioClinica, Inc.; Biogen; Bristol-Myers Squibb Company; CereSpir, Inc.; Cogstate; Eisai Inc.; Elan Pharmaceuticals, Inc.; Eli Lilly and Company; EuroImmun; F. Hoffmann-La Roche Ltd and its affiliated company Genentech, Inc.; Fujirebio; GE Healthcare; IXICO Ltd.; Janssen Alzheimer Immunotherapy Research \& Development, LLC.; Johnson \& Johnson Pharmaceutical Research \& Development LLC.; Lumosity; Lundbeck; Merck \& Co., Inc.; Meso Scale Diagnostics, LLC.;  NeuroRx Research; Neurotrack Technologies; Novartis Pharmaceuticals Corporation; Pfizer Inc.; Piramal Imaging; Servier; Takeda Pharmaceutical Company; and Transition Therapeutics. The Canadian Institutes of Health Research is providing funds to support ADNI clinical sites in Canada. Private sector contributions are facilitated by the Foundation for the National Institutes of Health (www.fnih.org). The grantee organization is the Northern California Institute for Research and Education, and the study is coordinated by the Alzheimer’s Therapeutic Research Institute at the University of Southern California. ADNI data are disseminated by the Laboratory for Neuro Imaging at the University of Southern California.  

\section{Funding}
This research did not receive any specific grant from funding agencies in the public, commercial, or not-for-profit sectors.

\end{document}